\documentclass[letterpaper, 10 pt, conference]{ieeeconf}  
\usepackage{CJKutf8}
\usepackage{graphicx}
\usepackage{capt-of}

\IEEEoverridecommandlockouts                              

\usepackage{graphics} 
\usepackage{epsfig} 
\usepackage{mathptmx} 
\usepackage{times} 
\usepackage{amsmath} 
\usepackage{amssymb}  
\usepackage{url}
\title{\LARGE \bf
ECHO-G: Embodied Co-speech Humanoid mOtion Generation
}

\author{%
Yizhao Li$^{2,4,*}$,
Pusen Gao$^{3,4,*}$,
Ming Wang$^{2}$,\\
Shaojie Shen$^{3}$,
Shuo Yang$^{4}$,
and Hao Xu$^{1,\dagger}$%
\thanks{$^{*}$Yizhao Li and Pusen Gao contributed equally
to this work.}%
\thanks{$^{\dagger}$Corresponding author: Hao Xu.}%
\thanks{$^{1}$Hao Xu is with Nanjing University.
Email: \texttt{xuhao3e8@nju.edu.cn}.}%
\thanks{$^{2}$Yizhao Li and Ming Wang are with
Beihang University.
Emails: \texttt{liyizhao@buaa.edu.cn},
\texttt{wangming@buaa.edu.cn}.}%
\thanks{$^{3}$Pusen Gao and Shaojie Shen are with
The Hong Kong University of Science and Technology.
Emails: \texttt{pgaoak@connect.ust.hk},
\texttt{eeshaojie@ust.hk}.}%
\thanks{$^{4}$Yizhao Li, Pusen Gao, and Shuo Yang
are with Mondo Robotics.
Email: \texttt{shuo.yang@mondorobotics.com}.}%
}

\usepackage{graphicx}     
\usepackage{booktabs}     
\usepackage{multirow}     
\usepackage[table]{xcolor}
\usepackage{makecell}
\usepackage{pifont}

\definecolor{TableGray}{gray}{0.93}
\definecolor{darkred}{RGB}{150,20,20}
\newcommand{\best}[1]{\textbf{#1}}

\newcommand{\second}[1]{\underline{#1}}
\newsavebox{\ECHOteaserbox}
\DeclareMathAlphabet{\mathcal}{OMS}{cmsy}{m}{n}
\makeatletter
\let\NAT@parse\undefined
\makeatother

\usepackage[unicode,hidelinks]{hyperref}
\hypersetup{
    pdftitle={ECHO-G: Embodied Co-speech Humanoid mOtion Generation},
    pdfauthor={Yizhao Li, Pusen Gao, Ming Wang, Shaojie Shen,
               Shuo Yang, Hao Xu}
}
\begin{document}

\begin{CJK*}{UTF8}{gbsn}

\begin{lrbox}{\ECHOteaserbox}
    \begin{minipage}{\textwidth}
        \centering
        \includegraphics[width=0.98\textwidth]{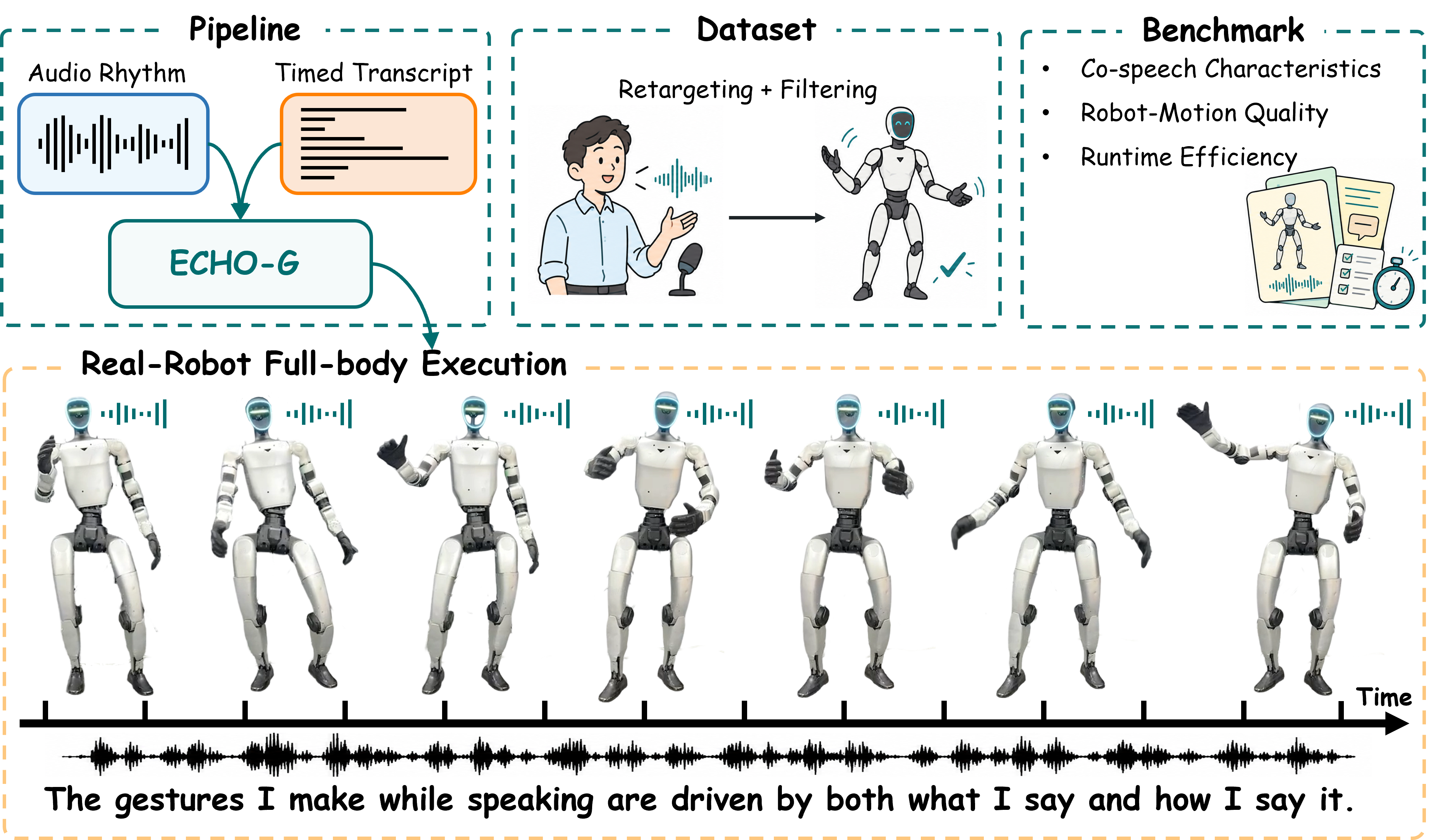}

        \captionof{figure}{
        \textbf{Overview of ECHO-G.}
        \emph{Top:} Speech audio and a timed transcript jointly
        condition full-body robot-motion generation.
        We publicly release a BEAT2-derived audio--text--robot
        dataset and a benchmark with evaluation code covering
        co-speech characteristics, robot-motion quality, and
        runtime efficiency.
        The dataset panel summarizes retargeting and quality filtering.
        \emph{Bottom:} Frames from a real-robot deployment experiment
        are shown in temporal order with the corresponding speech
        waveform and transcript.
        }
        \label{fig:teaser}
    \end{minipage}
\end{lrbox}

\AddToHook{cmd/@maketitle/after}{%
    \par\vspace{0.5em}
    \noindent\usebox{\ECHOteaserbox}\par
    \vspace{0.7em}
}

\maketitle
\thispagestyle{empty}
\pagestyle{empty}

\begin{abstract}
Generating full-body co-speech motion for humanoid robots
requires coordinating speech prosody, linguistic content,
and embodiment-specific motion.
To this end, we present ECHO-G, a framework jointly conditioned
on speech audio and timed transcripts.
Its Speech-Grounded Diffusion Transformer (SGDiT) combines
frame-aligned acoustic features with token-level linguistic
context, preserving their distinct granularities.
Trained with rectified flow matching, it models one-to-many
utterance--motion relationships directly in robot space.
To support training and evaluation, we introduce a BEAT2-derived
audio--text--robot dataset and a benchmark covering co-speech
characteristics, robot-motion quality, and runtime efficiency.
Comparative evaluation supports direct robot-space generation
over the evaluated human-motion generation and retargeting
pipelines, while modality ablations highlight the benefits
of joint audio--text conditioning.
We further demonstrate deployment on a physical humanoid robot.
A complementary video-rating study also favors joint
conditioning over the alternatives.
The dataset and training, inference, and evaluation code
are available through
\mbox{\href{https://echo-g-project.github.io/}{our project page}}.
\end{abstract}


\section{Introduction}
\label{sec:introduction}

In human communication, gestures complement spoken content,
convey emphasis, and organize the temporal structure of
speech~\cite{mcneill1992hand,kendon2004gesture}.
Inspired by this coordination, we aim to equip speaking
humanoids with body motion that reflects both how an
utterance is spoken and what it conveys, while respecting
the robot's embodiment.
We study full-body co-speech motion generation from the
audio and timed transcript of the robot's own utterance.

Human co-speech research provides a foundation for learning
speech--motion relationships. Representative methods explicitly
combine acoustic features with transcript-derived linguistic
features to model speech rhythm and content
~\cite{yoon2020speech,liu2024emage,liu2025gesturelsm}.
These methods primarily generate human-motion representations.
Extending speech-conditioned generation to humanoids requires
robot-specific motion references and an interface for their
physical execution.
Recent robot-oriented methods address the generation of
such references from speech. RoboGesture
~\cite{wang2026robogesture} studies audio-driven streaming
generation of upper-body and hand gestures, while
PhysDrift~\cite{liang2026physdrift} explores robot-native
generation with speech and text encoders.

These advances suggest a full-body robot co-speech generator
should combine densely sampled acoustic cues with token-level
linguistic content while preserving their distinct granularities.
The one-to-many relationship between utterances and gestures
~\cite{li2021audio2gestures} further motivates a generative
formulation, and real-robot deployment favors direct prediction
in robot space. In addition, existing public releases do not
consistently provide paired audio--text--robot training data
together with a benchmark spanning co-speech characteristics,
robot-motion quality, and runtime efficiency.

Motivated by these considerations, we present \textbf{ECHO-G},
a framework for full-body humanoid co-speech generation
from speech audio and timed transcripts
(Fig.~\ref{fig:teaser}).
Its \textbf{Speech-Grounded Diffusion Transformer (SGDiT)}
adds frame-aligned acoustic features to motion tokens and
retrieves token-level linguistic context through
global--local cross-attention, preserving the distinct
granularities of the two conditions.
Trained with rectified flow
matching~\cite{liu2022rectifiedflow}, SGDiT models the
one-to-many relationship between utterances and full-body
robot motion, enabling different motions to be sampled
for the same utterance.
A fixed pretrained whole-body motion tracker executes
the joint-position components of the generated references.

To support training and evaluation in humanoid co-speech
generation, we construct a BEAT2-derived robot-space dataset
through retargeting and embodiment-specific quality
filtering~\cite{liu2024emage,araujo2025retargeting}.
We publicly release the dataset and code for training,
inference, and evaluation, together with a benchmark covering
co-speech characteristics, robot-motion quality, and
runtime efficiency.
Pipeline comparisons and modality ablations assess the
generation-space and conditioning choices, while video
ratings and physical demonstrations provide complementary
perceptual and deployment evidence.

Our contributions are threefold:
\begin{itemize}

\item We present ECHO-G, a full-body humanoid co-speech
generation framework that jointly uses speech audio and timed
transcripts, and demonstrate its deployment on a physical
humanoid.

\item We develop SGDiT, a rectified-flow model combining
frame-aligned acoustic conditioning with global--local
transcript cross-attention for one-to-many full-body
robot-motion generation.

\item We release a BEAT2-derived dataset pairing speech audio
and timed transcripts with full-body robot motion, together with training, inference, and evaluation code.
The accompanying benchmark covers co-speech characteristics,
robot-motion quality, and runtime efficiency.

\end{itemize}

\section{Related Work}
\label{sec:related-work}
\begin{figure*}[!t]
      \centering
      \IfFileExists{fig/Arch-5.png}{
        \includegraphics[width=0.98\textwidth]{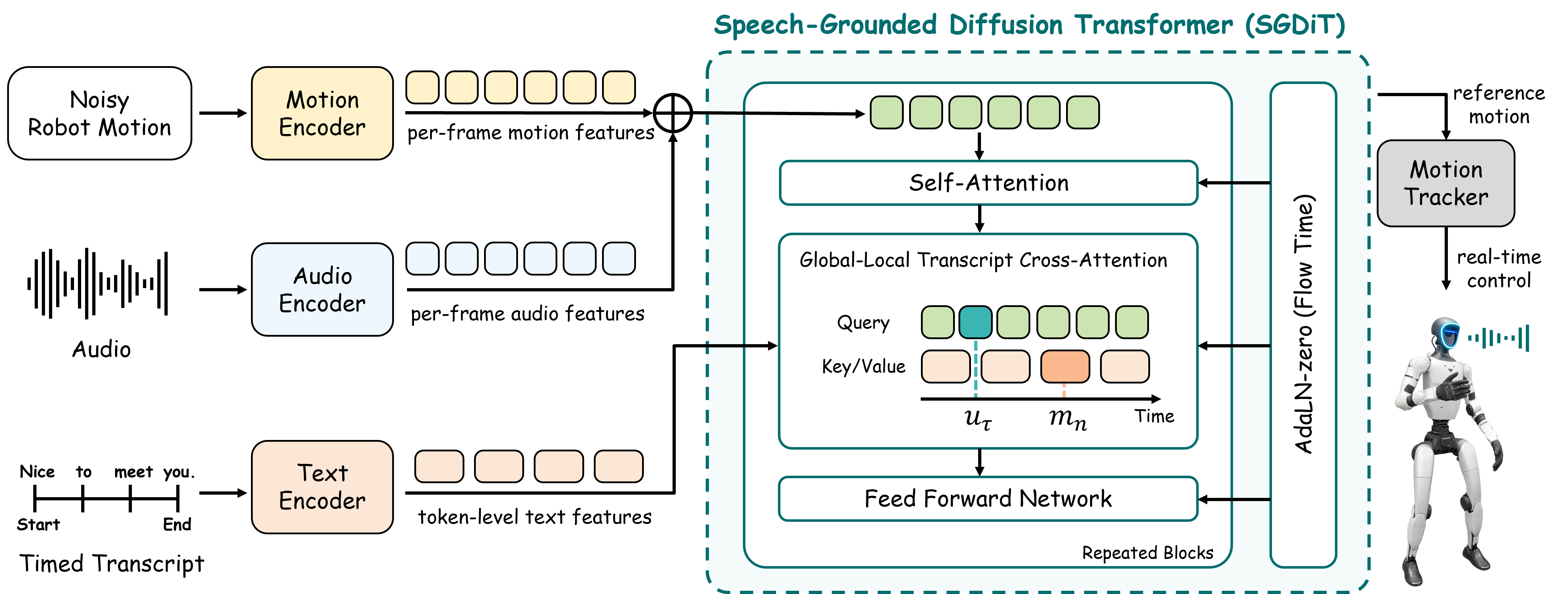}
      }{
        \fbox{\parbox{0.95\textwidth}{[[FIGURE: System overview. Speech audio and timed transcript are encoded, SGDiT generates robot motion, and the joint-position reference is tracked on the physical robot.]]}}
      }
        \caption{
        \textbf{SGDiT architecture and tracking interface.}
        Frame-aligned acoustic features are combined with noisy
        motion-frame features to form motion tokens.
        Contextual transcript embeddings provide a shared key--value
        memory for global and temporally biased text attention.
        Their attention distributions are fused within each
        transformer block before value aggregation and residual
        injection into the motion stream.
        Rectified-flow sampling produces robot-motion references,
        whose joint-position components are passed to a fixed
        whole-body motion tracker for execution.
        }
        \label{fig:system}
\end{figure*}

\subsection{Humanoid Whole-Body Motion Generation}

Recent whole-body tracking systems
~\cite{luo2025sonic,chen2026holomotion} enable humanoids
to execute diverse motion references.
Such references can be obtained by retargeting human motion,
as in GMR~\cite{araujo2025retargeting} and
OmniRetarget~\cite{yang2025omniretarget}, or generated from
language instructions, as in FRoM-W1~\cite{li2026w1} and
TextOp~\cite{textop}.
OMG~\cite{huang2026omg} further unifies language, audio,
and human-motion conditioning within a shared generative
framework.

Within this broader setting, robot co-speech generation
focuses on gestures accompanying spoken utterances.
Yoon et al.~\cite{yoon2019robots} generate transcript-conditioned
upper-body gestures and demonstrate execution on NAO.
RoboPerform~\cite{roboperform} generates humanoid motion from
speech audio using a generic text prompt rather than the
utterance transcript.
RoboGesture~\cite{wang2026robogesture} combines hierarchical
semantic--acoustic conditioning with streaming generation
of upper-body and hand gestures.
PhysDrift~\cite{liang2026physdrift} uses separate speech and
text encoders for one-step robot-native motion generation.
However, these approaches either omit utterance-specific
audio--text conditioning, focus on upper-body motion, or
do not fully specify the temporal organization of their
multimodal features.
ECHO-G combines frame-aligned acoustic features with
token-level transcript embeddings for full-body robot-space
generation, preserving their distinct granularities.
We also provide paired audio--text--robot data and code
for training, inference, and evaluation.

\subsection{Holistic Human Co-Speech Motion Generation}

BEAT~\cite{liu2022beat} provides multimodal speech--gesture
data and introduces CaMN for integrating audio, text,
and auxiliary conditions.
Building on BEAT2, EMAGE~\cite{liu2024emage} combines
adaptive content--rhythm fusion with masked gesture modeling
and compositional motion priors for holistic generation.
DiffSHEG~\cite{chen2024diffsheg} jointly generates expressions
and gestures through diffusion, while
GestureLSM~\cite{liu2025gesturelsm} combines flow matching,
latent shortcut learning, and spatiotemporal modeling of
body regions for efficient gesture generation.
These methods primarily synthesize human-motion representations.

Evaluation considers distributional fidelity, motion variation,
and speech--motion alignment.
Yoon et al.~\cite{yoon2020speech} introduced Fr\'echet Gesture
Distance (FGD), and EMAGE~\cite{liu2024emage} adopted
skeleton-aware features for distributional evaluation.
Audio2Gestures~\cite{li2021audio2gestures} examines motion
diversity and multimodality, while beat-alignment measures
assess temporal correspondence between motion and
audio~\cite{li2021ai,liu2024emage}.
Our benchmark adapts these evaluation dimensions to robot
motion and complements them with measures of robot-motion
quality and runtime efficiency.

\section{Method}
\label{sec:method}
\begin{figure*}[t]
      \centering
      \IfFileExists{fig/CA-6.png}{
        \includegraphics[width=0.98\textwidth]{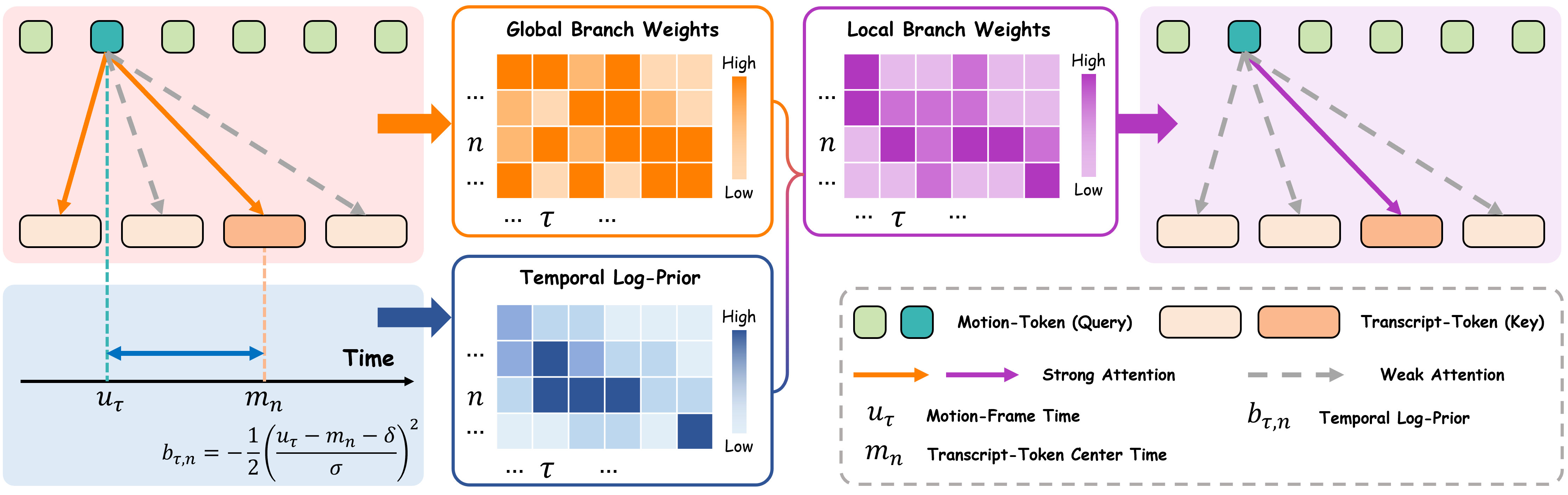}
      }{
        \fbox{\parbox{0.95\textwidth}{[[FIGURE: SGDiT architecture. Audio features condition motion frames, transcript tokens form cross-attention memory, word-time bias aligns motion frames with nearby words, and rectified flow predicts robot motion.]]}}
      }
        \caption{
        \textbf{Temporal conditioning in the local transcript-attention branch.}
        Orange and purple heatmaps show global and local attention
        weights, respectively.
        The blue heatmap represents the clipped Gaussian log-prior
        $\widetilde b_{\tau n}$ derived from motion-frame times
        $u_\tau$ and token-center times $m_n$.
        Multiplying global weights by the exponentiated log-prior
        and renormalizing yields the local weights.
        Matrices are transposed for display; colored and dashed
        arrows indicate stronger and weaker attention, respectively.
        }
        \label{fig:sgdit}
\end{figure*}

As shown in Fig.~\ref{fig:system}, ECHO-G generates full-body
robot-motion references from speech audio and timed transcripts.
SGDiT integrates acoustic and linguistic conditions within a
rectified-flow model, while a fixed whole-body motion tracker
executes the generated joint-position references.

\subsection{Problem Formulation and Motion Representation}
\label{subsec:problem}

Given speech audio $a$ and its word-timed transcript $y$,
ECHO-G models a conditional distribution over full-body
robot-motion sequences:
\begin{equation}
p_\theta(\mathbf R \mid a,y),
\label{eq:robot-motion}
\end{equation}
where $\theta$ denotes the generator parameters and
$\mathbf R=[\mathbf r_1,\ldots,\mathbf r_T]^\top
\in\mathbb R^{T\times D}$ contains $T$ motion frames.
For each frame $\tau=1,\ldots,T$, we use $D=39$ features:
\begin{equation}
\mathbf r_\tau
=
\bigl[
\boldsymbol{\rho}_\tau^\top,\,
\delta\psi_\tau,\,
(\mathbf v_\tau^{\mathrm{loc}})^\top,\,
\mathbf q_\tau^\top
\bigr]^\top,
\label{eq:robot-repr}
\end{equation}
where $\boldsymbol{\rho}_\tau\in\mathbb R^6$ encodes the base
orientation using a 6-D rotation representation,
$\delta\psi_\tau\in\mathbb R$ is the inter-frame yaw increment,
$\mathbf v_\tau^{\mathrm{loc}}\in\mathbb R^3$ is the base
linear velocity in a yaw-aligned local frame, and
$\mathbf q_\tau\in\mathbb R^{29}$ contains the robot's joint
angles in a fixed order.
Absolute root translation is omitted to make the learning
target invariant to global position offsets.

The generator operates on normalized motion features:
\begin{equation}
\mathbf x_\tau =
(\mathbf r_\tau-\boldsymbol{\mu})
\oslash\boldsymbol{\sigma},
\label{eq:motion-normalization}
\end{equation}
where $\boldsymbol{\mu},\boldsymbol{\sigma}\in\mathbb R^D$
are the feature-wise mean and standard deviation computed
from the training split, and $\oslash$ denotes element-wise
division.
We denote the normalized sequence by
$\mathbf X=[\mathbf x_1,\ldots,\mathbf x_T]^\top$.
Generated sequences are denormalized before evaluation
or execution.

\subsection{Audio--Text Conditioning}
\label{subsec:conditioning}

Acoustic features extracted by a frozen speech
encoder~\cite{baevski2020wav2vec} are linearly interpolated
to the $T$ motion frames.
The transcript text in $y$ is tokenized into $w_{1:N}$
and encoded by a frozen language
model~\cite{qwen2026qwen35}.
Separate layer normalization and learned affine projections
map the two feature sequences to
$\mathbf A\in\mathbb R^{T\times d}$ and
$\mathbf H\in\mathbb R^{N\times d}$, respectively,
where $d$ is the generator's hidden dimension and $N$
is the number of transcript tokens.
The projected conditions retain their frame-level and
token-level organization.

Using tokenizer character offsets, we derive approximate
token intervals $B=\{(s_n,e_n)\}_{n=1}^{N}$ from the
word-level timestamps, where $s_n$ and $e_n$ are the
associated start and end times.
Temporal conditioning uses token centers
\mbox{$m_n=(s_n+e_n)/2$} and motion-frame times
\mbox{$u_\tau=(\tau-1)/f$}, where $f$ is the motion frame rate.
Both times are measured from the utterance onset.
The combined condition is
\mbox{$c=(\mathbf A,\mathbf H,B)$}.

\subsection{Speech-Grounded Diffusion Transformer}
\label{subsec:motiondit}

SGDiT maps a noisy normalized motion sequence
$\mathbf X_t\in\mathbb R^{T\times D}$ and conditions $c$
to the flow velocity $v_\theta(\mathbf X_t,t,c)$.
Its transformer blocks combine temporal self-attention,
transcript cross-attention, and feed-forward processing.
Flow time $t\in[0,1]$ modulates the blocks through
adaptive layer normalization~\cite{peebles2023dit}.

\textbf{Acoustic conditioning.}
Each input motion token combines a projected motion frame,
its aligned acoustic condition, and a positional embedding:
\begin{equation}
\mathbf z_\tau
=
\mathbf W_r\mathbf x_{t,\tau}
+
\mathbf A_\tau
+
\mathbf p_\tau,
\label{eq:token-init}
\end{equation}
where $\mathbf x_{t,\tau}$ is frame $\tau$ of $\mathbf X_t$,
$\mathbf W_r\in\mathbb R^{d\times D}$ is a learned projection,
and $\mathbf p_\tau\in\mathbb R^d$ is a learned
frame-position embedding.
Temporal self-attention then exchanges information
bidirectionally across the acoustically conditioned
motion sequence.

\textbf{Transcript conditioning.}
Cross-attention retrieves linguistic context through two
paths over the same transcript features.
The global path provides content-based access to the full
token sequence, while the local path adds a preference for
temporally nearby tokens.
Updated motion features, after normalization and
flow-time modulation, provide queries $\mathbf Q_\tau$;
the projected transcript features $\mathbf H$ provide
keys $\mathbf K_n$ and values $\mathbf V_n$ shared by
both paths.
For one attention head, the content scores and global
attention are
\begin{equation}
\begin{aligned}
S_{\tau n}
&=
\gamma\,
\hat{\mathbf Q}_\tau^\top\hat{\mathbf K}_n,\\
\boldsymbol{\Pi}^{\mathrm g}
&=
\operatorname{maskedSoftmax}(\mathbf S),
\end{aligned}
\label{eq:qknorm-score}
\end{equation}
where hats denote L2-normalized queries and keys,
$\gamma$ is a bounded learned logit scale, and masked
softmax normalizes over non-padding transcript tokens.

The local path adds a Gaussian temporal prior to the shared
content scores, yielding the temporally reweighted attention
illustrated in Fig.~\ref{fig:sgdit}:
\begin{equation}
\begin{aligned}
b_{\tau n}
&=
-\frac{(u_\tau-m_n-\delta)^2}{2\sigma^2},\\[2pt]
\widetilde b_{\tau n}
&=
\max(b_{\tau n},-\kappa),\\[2pt]
\boldsymbol{\Pi}^{\mathrm l}
&=
\operatorname{maskedSoftmax}
(\mathbf S+\widetilde{\mathbf b}),
\end{aligned}
\label{eq:timed-xa}
\end{equation}
where $\sigma$ and $\delta$ control the temporal width
and offset, and $\kappa$ bounds the log penalty.
Both paths use the same token padding mask.

The two distributions are combined using temporal support:
\begin{equation}
\begin{aligned}
g_\tau
&=
\alpha\max_{n\in I}\exp(b_{\tau n}),\\
\Pi_{\tau n}
&=
(1-g_\tau)\Pi^{\mathrm g}_{\tau n}
+
g_\tau\Pi^{\mathrm l}_{\tau n},
\end{aligned}
\label{eq:text-fusion}
\end{equation}
where $I$ contains the non-padding token indices and
$\alpha$ is the learned base mixing coefficient.
Support uses the unclipped prior, reducing the local
contribution when the frame is distant from all
offset-adjusted token centers.
The mixed weights aggregate the shared values into a
text-conditioned update, which is projected and added
to the motion features through a gated residual connection.
A linear output head produces the final
$T\times D$ flow-velocity prediction.

\subsection{Training Objective}
\label{subsec:training}

We train SGDiT with rectified flow matching
~\cite{liu2022rectifiedflow,lipman2023flowmatching}.
For a normalized motion--condition pair $(\mathbf X,c)$,
we sample standard Gaussian noise
$\boldsymbol{\epsilon}\in\mathbb R^{T\times D}$ and
a sequence-level flow time $t\sim\mathcal U(0,1)$.
The interpolated motion and target velocity are
\begin{equation}
\mathbf X_t=t\mathbf X+(1-t)\boldsymbol{\epsilon},
\qquad
\mathbf V^\star=\mathbf X-\boldsymbol{\epsilon}.
\label{eq:flow-target}
\end{equation}
The predicted velocity
$\hat{\mathbf V}=v_\theta(\mathbf X_t,t,c)$
is supervised by matching the target flow and its
adjacent-frame differences:
\begin{equation}
\begin{aligned}
\mathcal{L}_{\mathrm{flow}}
&=
\mathbb{E}\bigl[
\operatorname{MSE}
(\hat{\mathbf V},\mathbf V^\star)
\bigr],\\[2pt]
\mathcal{L}_{\mathrm{temp}}
&=
\mathbb{E}\bigl[
\operatorname{MSE}
(\Delta_\tau\hat{\mathbf V},
 \Delta_\tau\mathbf V^\star)
\bigr],\\[2pt]
\mathcal{L}_{\mathrm{gen}}
&=
\mathcal{L}_{\mathrm{flow}}
+
\lambda_{\mathrm{temp}}\mathcal{L}_{\mathrm{temp}}.
\end{aligned}
\label{eq:gen-loss}
\end{equation}
Here $\Delta_\tau$ denotes first differences along the
motion-frame axis, and $\lambda_{\mathrm{temp}}$ weights
the temporal term.
MSE is averaged over feature dimensions and valid frames;
the temporal term uses only adjacent pairs of valid frames.

\begin{figure*}[!t]
    \centering
    \includegraphics[width=\textwidth]{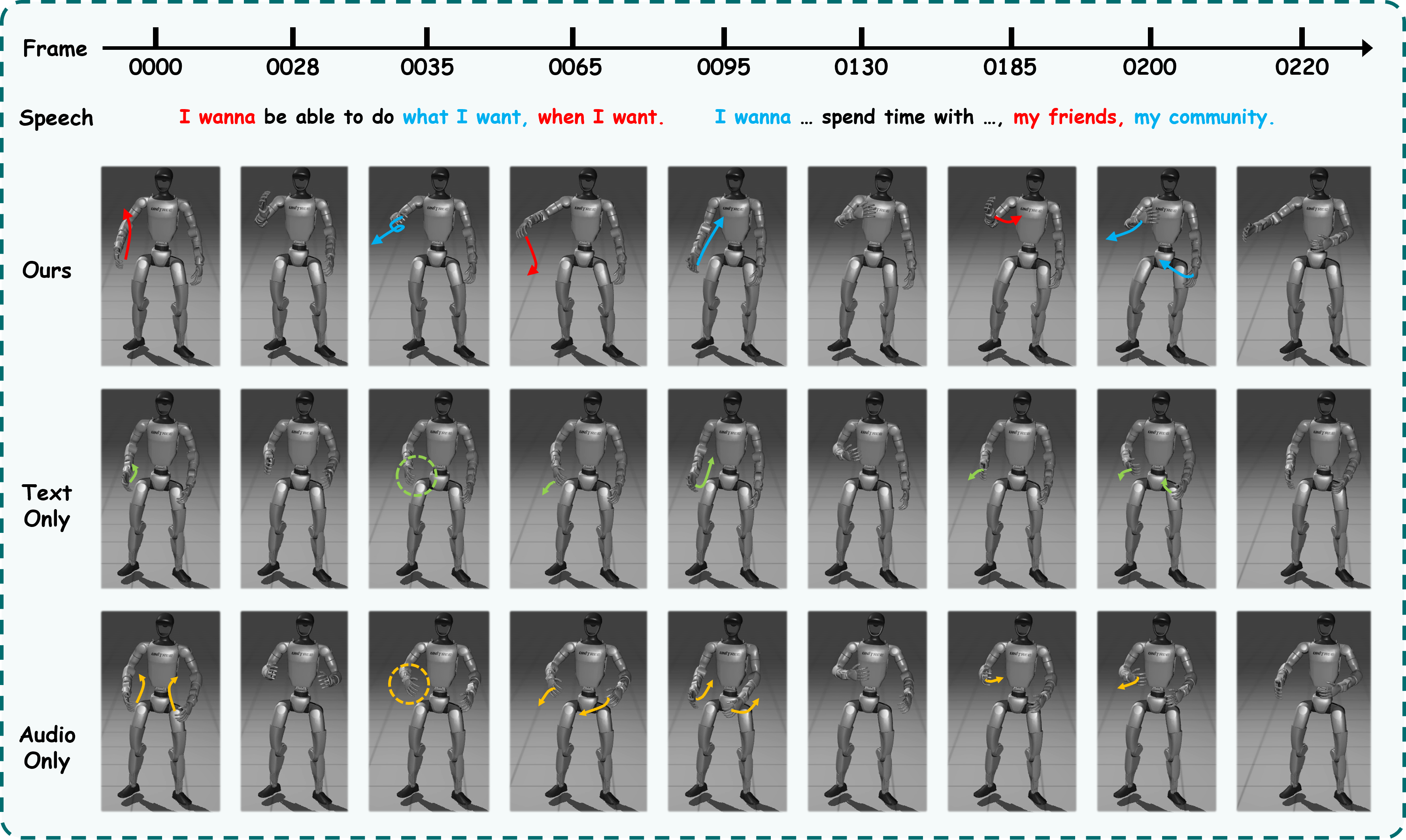}
    \caption{
    \textbf{Conditioning comparison on an utterance outside BEAT2.}
    Rows show joint audio--text conditioning (Ours), text-only,
    and audio-only outputs at matched frame indices.
    In the selected frames, joint conditioning exhibits broader
    arm extensions, whereas the unimodal outputs generally keep
    the hands closer to the torso.
    Colored arrows and circles highlight selected arm and
    hand movements.
    }
    \label{fig:modality_qualitative}
\end{figure*}
\subsection{Inference and Tracking}
\label{subsec:tracking}

At inference, the output length $T$ is specified by the
input clip duration at the motion frame rate.
We initialize $\mathbf X^{(0)}\in\mathbb R^{T\times D}$
with standard Gaussian noise and integrate the learned
velocity field from flow time $0$ to $1$, keeping $c$ fixed.
Using $K$ uniform Euler steps, we update
\begin{equation}
\begin{aligned}
t_k &= \frac{k}{K},\\
\mathbf X^{(k+1)}
&=
\mathbf X^{(k)}
+
\frac{1}{K}
v_\theta(\mathbf X^{(k)},t_k,c),
\end{aligned}
\label{eq:euler-sampling}
\end{equation}
for $k=0,\ldots,K-1$.
The final state $\hat{\mathbf X}=\mathbf X^{(K)}$ is
converted to robot-motion references by inverting
the normalization in~\eqref{eq:motion-normalization}.
Their joint-angle components are supplied to the fixed
SONIC motion tracker~\cite{luo2025sonic} as joint-position
references and executed alongside speech playback.

\section{Experiments and Results}

\subsection{Experimental Setup}

\subsubsection{Datasets and Preprocessing}

We construct a robot-space co-speech dataset from
BEAT2~\cite{liu2024emage}. The original long-form recordings are
segmented into 22,192 utterance-level clips at speech pauses
using word-level forced alignment, retaining the corresponding
audio, transcript, SMPL-X motion, and word timestamps.
Each motion sequence is retargeted to the 29-DoF Unitree G1
using GMR~\cite{araujo2025retargeting}. The retargeted motions are
converted to a unified Z-up coordinate system and foot-ground
aligned using a clip-wise vertical root offset, followed by
recomputation of forward kinematics and motion derivatives.
We further canonicalize each sequence by removing its initial
global yaw while preserving subsequent root dynamics. Robot
motions remain at the native 30~fps throughout processing.

We then filter the resulting speech--robot pairs using both
robot-motion quality checks and cross-modal consistency checks.
For robot-motion quality, we discard clips that violate criteria
on foot contact, self-collision, joint continuity and limits,
smoothness, or severe high-frequency motion artifacts. We also
retain only pairs whose audio and motion durations differ by
at most one motion frame.

We adopt a speaker-held-out split, holding out three English
speakers for validation and using the remaining speakers
for training.
After filtering, the final dataset contains 14,987 training
clips and 3,242 validation clips.

\subsubsection{Evaluation Metrics}

All methods use the same held-out split as the candidate input set,
with eligibility determined by their native input and output-length
requirements. Prediction--reference comparisons use the common
temporal prefix of each eligible pair.

\textbf{Co-Speech Motion Characteristics.}
Fr\'echet Gesture Distance (FGD)~\cite{yoon2020speech,liu2024emage}
measures distributional discrepancy using a shared skeleton-aware
G1 joint-motion encoder. Div, MM, and BA are computed from
forward-kinematic body positions with fixed base rotation and
translation. Diversity (Div) measures the frame-wise L1 deviation
from each sequence's temporal mean pose. For stochastic generators,
Multimodality (MM)~\cite{li2021audio2gestures} averages pairwise L1
distances among 20 samples generated for each input condition.
Beat Alignment (BA)~\cite{li2021ai,liu2024emage} matches speech
onsets to the nearest detected upper-body motion beats.
We report absolute gaps to the matched reference statistics
for Div and BA.

\textbf{Robot Motion Quality.}
\begin{figure*}[!t]
    \centering
    \includegraphics[width=\textwidth]{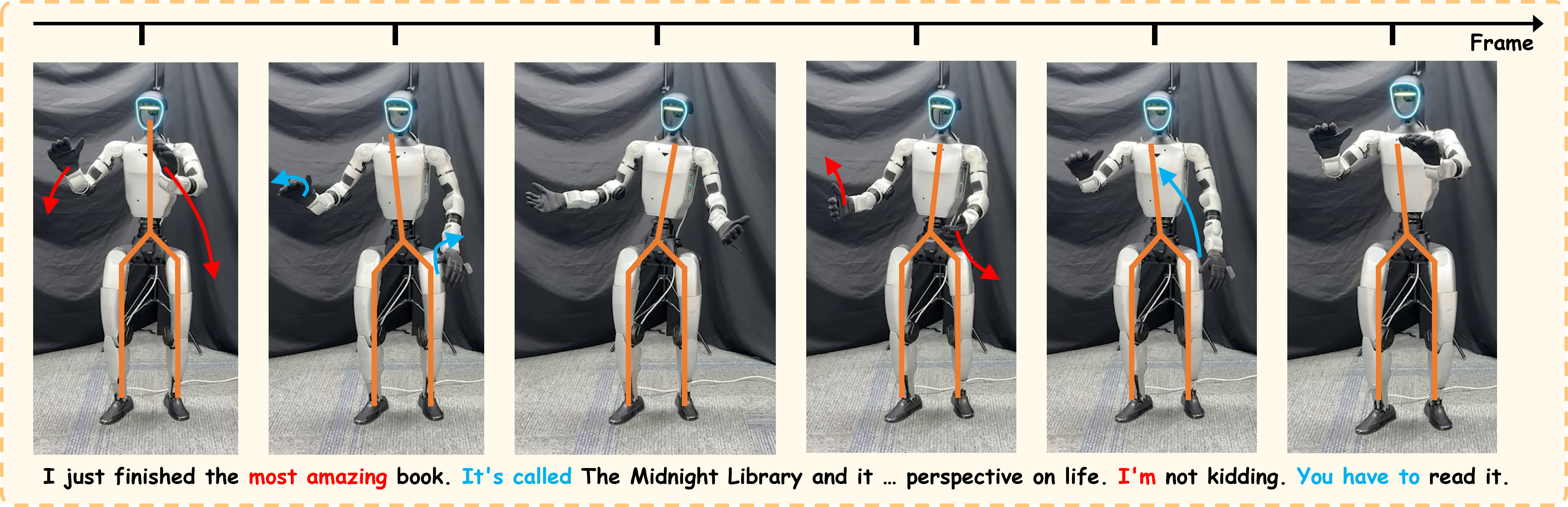}
    \caption{
    \textbf{Real-robot execution on an utterance outside BEAT2.}
    ECHO-G's predicted joint angles are supplied as joint-position
    references to a fixed SONIC motion tracker while the corresponding
    speech is played.
    Frames progress from left to right through the book-recommendation
    utterance shown below, illustrating changes in arm extension,
    hand height, and torso posture.
    Colored arrows highlight selected arm movements, while the
    orange skeletal overlays outline the body configuration.
    }
    \label{fig:robot_qualitative}
\end{figure*}
Body jerk~\cite{huang2026omg} is estimated from third-order
finite differences of reconstructed world-space body positions,
scaled by the cube of the frame rate. We average jerk magnitudes
over all valid frame--body pairs and report the absolute gap
between generated and reference means
($\Delta$Jerk)~\cite{Fang_2026_CVPR}.
Foot-ground error measures the vertical distance of the lowest
sole-proxy surface from the ground plane. Contact sliding speed
measures the maximum horizontal sole-point speed per foot,
averaged over detected contact intervals.

\textbf{Efficiency.}
End-to-end inference time covers input audio and aligned-transcript
reading, online feature encoding, motion generation, and
GMR or VAE-based motion mapping when applicable, ending at
the robot-motion reference.
We report total processing time divided by the total number of output
frames. Peak RAM increase is the maximum request-window system
memory usage above the corresponding pre-load idle baseline.

\subsubsection{Implementation Details}

We use frozen wav2vec~2.0 large XLSR and Qwen3.5-4B models
to obtain 1024-D acoustic features and 2560-D contextual
token embeddings, respectively.
SGDiT contains 12 transformer blocks with a hidden dimension
of 768, 8 attention heads, a feed-forward dimension of 2048,
and learned temporal positional embeddings.
The attention parameters $\gamma,\sigma,\delta,\alpha$
are learned separately for each layer and head.
Attention parameters are constrained to
$\gamma\in[1,16]$, $\sigma\in[0.25,2]$~s,
$\delta\in[-0.5,0.5]$~s, and $\alpha\in[0,0.5]$
using sigmoid/tanh mappings, with log-prior clipping
threshold $\kappa=12$.

We train for 63,000 optimizer steps on NVIDIA RTX 4090
hardware using FP32 computation.
AdamW uses an initial learning rate of $3\times10^{-4}$
with cosine decay and no warm-up, weight decay of $10^{-4}$,
an effective batch size of 48, and gradient clipping at 1.0.
We set $\lambda_{\mathrm{temp}}=0.5$ and jointly drop
the acoustic and text encoder features together with
the time-distance inputs with probability 0.1.
The exponential moving average (EMA) decay is 0.999.

For physical deployment, motion generation runs on a Jetson
AGX Orin, which is also used for the efficiency evaluation
of all compared methods.
Inference uses EMA weights from the checkpoint with the
lowest EMA validation loss.
Sampling uses eight Euler steps with classifier-free
guidance (CFG) scale 1.0.
We generate at most 600 motion frames at 30~fps and limit
transcripts to 256 tokenizer tokens. Efficiency is measured with batch size one after warm-up.

\subsubsection{Baselines and Ablations}

We use the publicly released pretrained checkpoints of
EMAGE~\cite{liu2024emage} and
GestureLSM~\cite{liu2025gesturelsm}, and retarget their
generated human motions to G1 using
GMR~\cite{araujo2025retargeting}.
We additionally construct Human-Retargeted using the same
audio--text conditioning design, backbone configuration,
data split, and optimization settings as Ours, but generate
normalized 136-D human motion.
After denormalization using human-motion training statistics,
a pretrained VAE-based mapping converts the samples to
39-D robot references.
This mapping remains frozen during human-motion generator
training.

Audio-only and Text-only are trained separately in robot
space using only the indicated modality.
Both variants share the data split, backbone configuration,
and optimization settings with Ours. Text-only retains the supplied clip duration and word timings
but does not use acoustic features.

\subsection{Quantitative Results}

\begin{table*}[!t]
\caption{
Comparison of direct robot-space generation with human-motion
generation followed by retargeting or learned mapping on the
BEAT2 speaker-held-out evaluation data.
$\Delta$Div, $\Delta$BA, and $\Delta$Jerk denote absolute deviations
from matched ground-truth statistics over each method's evaluated
motion range. Best and second-best results are shown in
\textbf{bold} and \underline{underlined} text, respectively.
}
\label{tab:a2m_main}

\centering
\footnotesize
\setlength{\tabcolsep}{2.2pt}
\renewcommand{\arraystretch}{1.10}

\begin{tabular*}{\textwidth}{
@{\extracolsep{\fill}}
lccccccccc
@{}
}
\toprule

\multirow{2}{*}{Method}
& \multicolumn{4}{c}{Co-Speech Motion Characteristics}
& \multicolumn{3}{c}{Robot Motion Quality}
& \multicolumn{2}{c}{Efficiency} \\

\cmidrule(lr){2-5}
\cmidrule(lr){6-8}
\cmidrule(lr){9-10}

& FGD $\downarrow$
& $\Delta$Div $\downarrow$
& MM $\uparrow$
& $\Delta$BA $\downarrow$
& $\Delta$Jerk $\downarrow$
& \makecell{Foot Err.\\(m) $\downarrow$}
& \makecell{C-Slide\\(m/s) $\downarrow$}
& \makecell{E2E Time\\(ms/frame) $\downarrow$}
& \makecell{Peak RAM $\Delta$\\(MB) $\downarrow$} \\

\midrule

EMAGE+GMR
& 4.976
& 0.749
& 0
& 0.172
& 26.951
& 0.013
& 0.163
& 21.3
& \best{2661} \\

GestureLSM+GMR
& 5.008
& 0.561
& 1.016
& \second{0.158}
& 24.444
& 0.010
& 0.169
& 20.5
& \second{3804} \\

Human-Retargeted
& \second{4.725}
& \second{0.408}
& \second{1.498}
& 0.161
& \best{8.001}
& \best{0.003}
& \best{0.050}
& \second{6.56}
& 18714 \\

\midrule

\rowcolor{TableGray}
\textbf{Ours}
& \best{2.278}
& \best{0.320}
& \best{1.786}
& \best{0.063}
& \second{9.039}
& \second{0.008}
& \second{0.052}
& \best{5.96}
& 18542 \\

\bottomrule
\end{tabular*}
\end{table*}
\begin{table*}[!t]
\caption{
Ablation of conditioning modalities for direct robot-space
generation on the BEAT2 speaker-held-out evaluation data.
$\Delta$Div, $\Delta$BA, and $\Delta$Jerk denote absolute deviations
from the shared ground-truth statistics.
Best and second-best results are shown in \textbf{bold} and
\underline{underlined} text, respectively.
Ties at the displayed precision receive identical highlighting.
}
\label{tab:comparison_ablation}

\centering
\footnotesize
\setlength{\tabcolsep}{3.5pt}
\renewcommand{\arraystretch}{1.10}

\begin{tabular*}{\textwidth}{
@{\extracolsep{\fill}}
lccccccc
@{}
}
\toprule

Method
& FGD $\downarrow$
& $\Delta$Div $\downarrow$
& MM $\uparrow$
& $\Delta$BA $\downarrow$
& $\Delta$Jerk $\downarrow$
& \makecell{Foot Err.\\(m) $\downarrow$}
& \makecell{C-Slide\\(m/s) $\downarrow$} \\

\midrule

Audio-only
& \second{2.360}
& \second{0.360}
& \second{1.702}
& \second{0.081}
& \best{4.497}
& \best{0.008}
& \best{0.038} \\

Text-only
& 2.436
& 0.429
& 1.681
& 0.113
& \second{5.940}
& \second{0.009}
& \second{0.049} \\

\midrule

\rowcolor{TableGray}
\textbf{Ours}
& \best{2.278}
& \best{0.320}
& \best{1.786}
& \best{0.063}
& 9.039
& \best{0.008}
& 0.052 \\

\bottomrule
\end{tabular*}
\end{table*}

Table~\ref{tab:a2m_main} compares direct robot-space generation
with human-motion generation followed by retargeting or learned
mapping. ECHO-G achieves the best results on all four co-speech
metrics and the lowest processing time per output frame.
It also improves all three robot-motion quality metrics over
EMAGE+GMR and GestureLSM+GMR.
Human-Retargeted yields a smaller Jerk gap, lower foot-ground
error, and less contact sliding.
These results support direct robot-space generation for
co-speech modeling with reduced processing time.

Table~\ref{tab:comparison_ablation} evaluates conditioning
modalities within the robot-space generator.
Joint audio--text conditioning achieves the lowest FGD,
$\Delta$Div, and $\Delta$BA and the highest MM, outperforming
both unimodal variants on the reported co-speech metrics.
Audio-only yields the smallest Jerk gap and contact sliding
speed, and matches joint conditioning in foot-ground error
at the displayed precision.
These results support combining acoustic and linguistic
information to improve the evaluated co-speech characteristics.

\subsection{Qualitative Results}

We visualize motions generated from independently prepared
audio--text utterances outside BEAT2.
These examples provide qualitative evidence of generalization
to speech inputs beyond the source dataset.
In Fig.~\ref{fig:modality_qualitative}, audio-only conditioning
produces rhythm-responsive motion, but gestures around
semantically salient phrases remain small and less clearly
related to the spoken content.
Text-only conditioning produces content-related gestures,
but their timing is less consistently aligned with the audio.
Joint conditioning combines speech-responsive timing with
more expansive, content-related gestures and fluid transitions
in this example.

\subsection{Real-Robot Deployment}

We conduct real-robot experiments on a Unitree G1, executing
the generated joint-position references through a fixed
SONIC motion tracker~\cite{luo2025sonic} while playing the
corresponding speech audio.
Fig.~\ref{fig:robot_qualitative} shows a representative trial
using an independently prepared utterance outside BEAT2,
with the robot accompanying its speech with generated
body movements.
Videos of additional real-robot trials are provided on the
\href{https://echo-g-project.github.io/}{project page}.

\subsection{User Study}
\begin{table}[!t]
\caption{
Mean user-study ratings from 45 participants on a five-point scale.
Higher is better. Unimodal scores are pooled as described in the text.
Best and second-best means are shown in \textbf{bold} and
\underline{underlined} text, respectively.
}
\label{tab:user_study}

\centering
\footnotesize
\setlength{\tabcolsep}{3.0pt}
\renewcommand{\arraystretch}{1.10}

\begin{tabular}{lcccc}
\toprule

Method
& Overall
& \makecell{Human-\\likeness}
& \makecell{Rhythm\\Matching}
& \makecell{Motion\\Quality} \\

\midrule

EMAGE+GMR
& 1.80
& 1.87
& 1.95
& 1.59 \\

Human-Retargeted
& 2.65
& 2.67
& 2.38
& 2.90 \\

Unimodal (pooled)
& \second{3.05}
& \second{3.01}
& \second{3.13}
& \second{3.01} \\

\rowcolor{TableGray}
\textbf{Ours}
& \best{3.49}
& \best{3.45}
& \best{3.31}
& \best{3.70} \\

\bottomrule
\end{tabular}

\end{table}

We conducted a video-rating study with 45 participants using
G1 kinematic renderings in MuJoCo. Each participant completed
nine trials, with three randomly selected from each of three
criterion-specific pools comprising 33 utterances in total.
The criteria were human-likeness, rhythm matching, and motion
quality. Each trial presented four videos of the same utterance
under matched rendering conditions, with hidden method identities
and randomized positions. Participants rated each video on a
five-point scale.

Each trial compared ECHO-G, Human-Retargeted, EMAGE+GMR, and
an audio-only or text-only variant. Unimodal ratings
were pooled over the observed trial allocation. Overall denotes the
equally weighted mean of the three criterion scores.
As shown in Table~\ref{tab:user_study}, joint audio--text
conditioning receives the highest overall mean rating and the
highest mean ratings across all three criteria, providing
perceptual support for our method.

\section{Discussion and Limitations}

The experimental results support direct robot-space modeling
and joint audio--text conditioning for humanoid co-speech
generation.
Benchmark comparisons show the benefits of these choices
for the reported co-speech characteristics, with the direct
generation pipeline also requiring less processing time.
Qualitative examples, user ratings, and physical demonstrations
provide complementary evidence of expressive gestures,
perceived quality, and robot execution.
Further improvement is needed to better reconcile expressive
behavior with robot-motion consistency.

Several limitations remain.
First, the gains in co-speech characteristics do not consistently
translate into smaller Jerk gaps or better foot-contact measures,
leaving room to improve expressive behavior and robot-motion
consistency together.
Second, the model learns broad speech--gesture associations,
with limited training examples of explicit deictic or
instructional gestures.
This may constrain instruction-aware gesture generation
when an utterance calls for a specific semantic motion.
Third, the current system requires complete speech audio
and timed transcripts and does not yet support causal
streaming generation.

\section{Conclusion}

We presented ECHO-G for full-body humanoid co-speech generation
from speech audio and timed transcripts. SGDiT combines frame-aligned acoustic conditioning with
global--local transcript cross-attention to generate
robot-motion references.
Quantitative evaluation, a video-rating study, and physical
demonstrations provide complementary evidence for the framework.
The released dataset, benchmark, and code support reproducible
research on humanoid co-speech generation. Future work will focus on jointly improving gesture
expressiveness and robot-motion consistency, enriching
training data for instruction-aware semantic gestures,
and extending the framework to causal streaming generation.


\bibliographystyle{IEEEtran}
\bibliography{references}

\end{CJK*}
\end{document}